\documentclass[10pt,journal,compsoc]{IEEEtran}
\usepackage[pagebackref=true, breaklinks=true, colorlinks, bookmarks=false]{hyperref}
 
\usepackage{graphicx} 
\usepackage{siunitx}
\usepackage[ruled,vlined]{algorithm2e}
\usepackage{algpseudocode}
\usepackage{amsmath}
\usepackage{amsfonts}
\usepackage{amssymb}
\usepackage{url} 
\usepackage{xcolor}
\usepackage{cite}
\usepackage[labelfont=bf,labelsep=space]{caption}
\usepackage{subcaption}
\usepackage{upgreek}

\title{Event Camera Based Real-Time Detection and Tracking of Indoor Ground Robots}
\author{Himanshu~Patel*, Craig~Iaboni*, Deepan~Lobo, Ji-won~Choi, Pramod~Abichandani

\IEEEcompsocitemizethanks{\IEEEcompsocthanksitem *Equal contribution. 

The authors are with the Robotics and Data Laboratory (RADLab) and Departments
of Electrical and Computer Engineering and Computer Science, New Jersey Institute of Technology, Newark,
NJ. \protect\\
E-mail: pva23@njit.edu
}
}

\date{June 2020}

\begin{document}

\IEEEtitleabstractindextext{%
\begin{abstract}
This paper presents a real-time method to detect and track multiple mobile ground robots using event cameras. The method uses density-based spatial clustering of applications with noise (DBSCAN) to detect the robots and a single k-dimensional ($k - d$) tree to accurately keep track of them as they move in an indoor arena. Robust detections and tracks are maintained in the face of event camera noise and lack of events (due to robots moving slowly or stopping). An off-the-shelf RGB camera-based tracking system was used to provide ground truth.  Experiments including up to 4 robots are performed to study the effect of i) varying DBSCAN parameters, ii) the event accumulation time, iii) the number of robots in the arena, iv) the speed of the robots, and v) variation in ambient light conditions on the detection and tracking performance. The experimental results showed 100\% detection and tracking fidelity in the face of event camera noise and robots stopping for tests involving up to 3 robots (and upwards of 93\% for 4 robots). When the lighting conditions were varied, a graceful degradation in detection and tracking fidelity was observed.
\end{abstract}

}

\maketitle

\section{Introduction}

The commercial availability of dynamic vision sensor (DVS) based cameras, also known as event cameras, has provided researchers and practitioners with an attractive modality for high-speed computer vision applications. By recording a change in light intensities asynchronously, event-based cameras offer several advantages over their frame-based camera counterparts to provide high speed vision, low perception latency, and relatively low power requirements \cite{liu2019event}.The earliest event-based systems were designed in 1986 at Caltech \cite{mahowald1994silicon} and have recently been the focus of significant commercial development by companies such as Prophesee, iniVation, Samsung, Insightness \cite{2008lichsteiner, 2011posch, 2014brandli, 2017son}.

Multiple ground robots based cooperative system are widely used in indoor applications such as warehouse automation, surveillance and security, and payload transportation \cite{amazonRobotics, festoRobotics, 6riverRobotics}. The task of detecting robots in indoor environments, estimating their pose with respect to the environment, and tracking their motion in real-time is crucial to developing of autonomous systems \cite{Kanellakis2017, Desouza2002, Shengyong2011}. Developments in this field have been made using a combination of traditional frame-based cameras and inertial measurements of the robot \cite{Kanellakis2017, Desouza2002, Shengyong2011, Robin2016MultirobotTD}. Tracking systems with the traditional modality of frame-based cameras suffer from frame rate limitations and have trouble with motion blur and dynamic range. Another tracking modality is an infrared camera-based motion capture systems such as OptiTrack and Vicon \cite{VICON, OptiTrack}. While these infrared systems provide fast and high-quality tracking data, they come at a significant price-tag, require multiple infrared cameras, are marker-based, and need powerful computing infrastructure for data acquisition and processing. 
By contrast, an event-camera negates the need to read an entire frame of image data since it only provides intensity information (positive/negative polarity) at a given location at any time, does not require markers on objects of interest, and can be operated at relatively lower computational complexity and power \cite{gallego2020}. Stationary event cameras do not require background modeling as moving objects can be easily detected against a static background. This significantly improves data throughput and processing, making event cameras beneficial for robotics applications \cite{gallego2020}. 

\begin{figure}[t!]
    \centering
            \includegraphics[scale = 0.08]{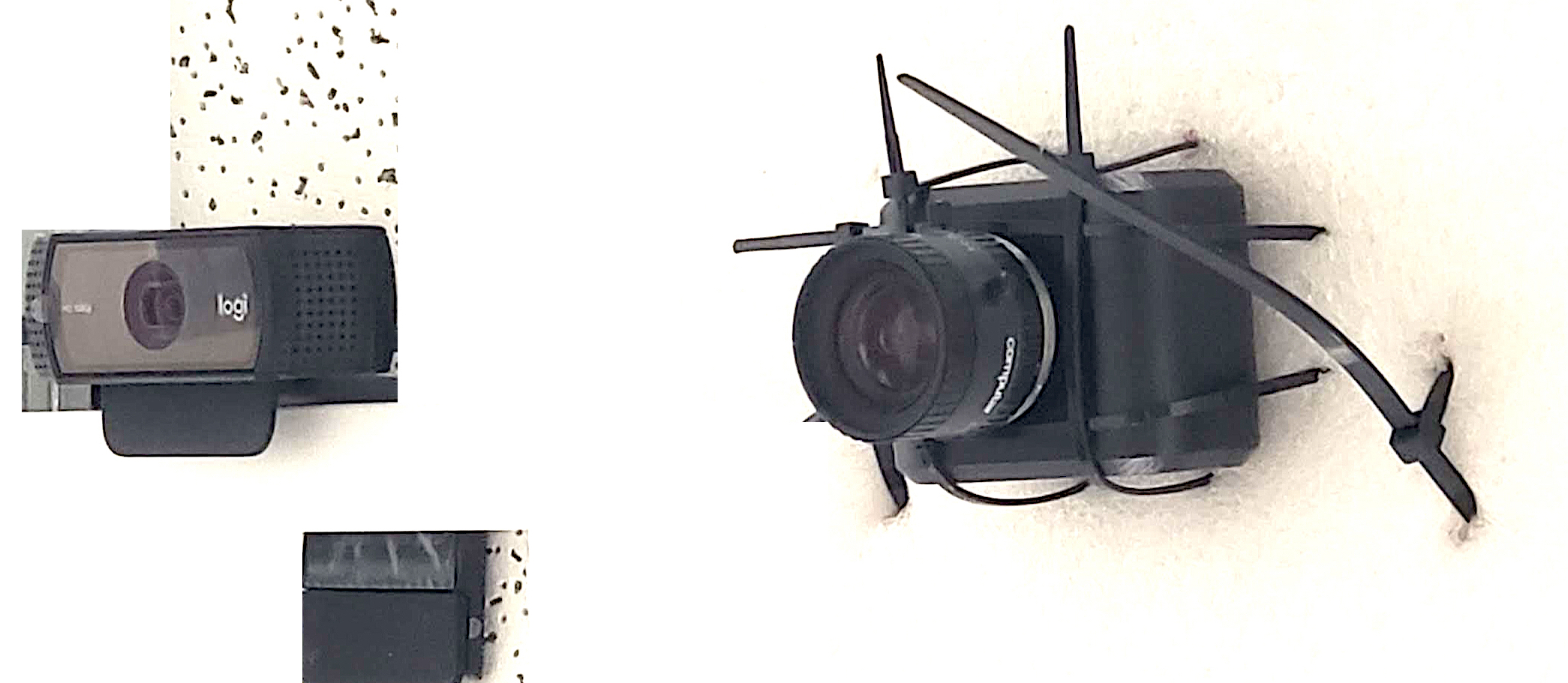}
    \includegraphics[scale=0.32]{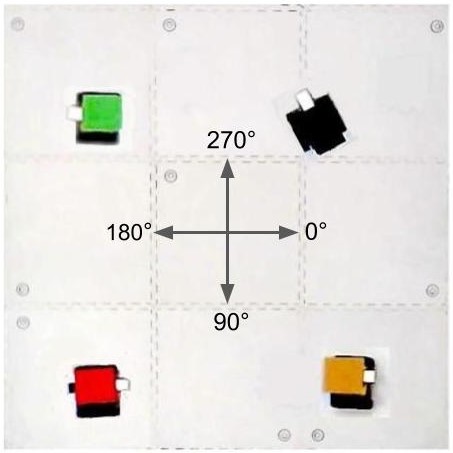}
    \includegraphics[scale = 0.3]{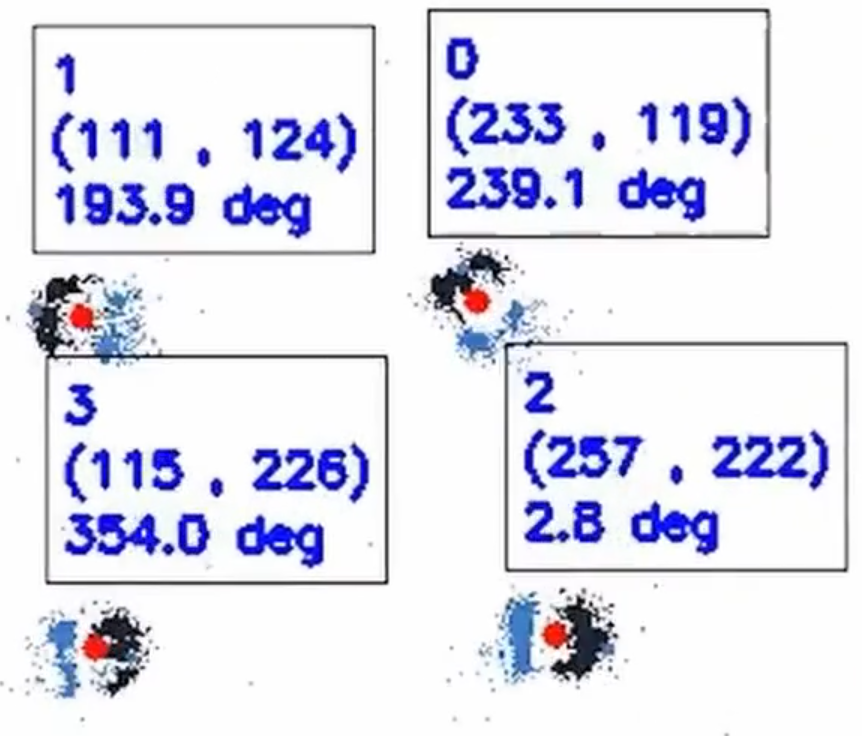}
    \caption{Top: The Prophesee Gen3S VGA-CD event camera was mounted on the ceiling facing downward. A Logitech webcam was mounted next to the event camera to provide ground truth data. Bottom left: Multi-robot mission (top view) with 4 ground robots. The robots moved in a 183 cm $\times$ 183 cm arena on white-colored foam mats. Bottom right: Event data visualization for 4 ground robots and annotations representing the detection and tracking results.}
    \label{hero_image}
\end{figure}

The benefits mentioned above form our primary motivation for developing a real-time detection and tracking method using an event-based camera for indoor robot operations. The event camera used in this study featured the Prophesee Gen3S VGA-CD dynamic vision sensor. 

The method uses density-based spatial clustering of applications with noise (DBSCAN) to detect the robots and a single k-dimensional ($k - d$) tree to accurately keep track of them as they move in an indoor arena \cite{Ester96adensity-based, Bentley1975}. DBSCAN is a powerful and popular clustering algorithm first proposed in 1996 and has found use in a plethora of data-driven applications \cite{Ester96adensity-based}.  The main idea is that a neighborhood of a cluster of points should have a minimum number of points in a given radius. A point is considered as part of a cluster as long as it has a minimum number of points ($minPts$) in its neighborhood of radius $eps$. DBSCAN is an effective clustering algorithm, especially when the clusters are arbitrarily shaped and noisy \cite{schubert2017dbscan}. In recent years, there have been debates about its effectiveness in cases with 3 or higher dimensions, and competing methods have been proposed \cite{gan2015dbscan}. However, the original authors of DBSCAN have shown that with effective indexes and reasonably chosen parameter values, DBSCAN performs competitively in higher dimensions \cite{Schubert2020}. 

The $k-d$ tree is a multi-dimensional binary search tree used widely to store information that is retrieved by associative searches \cite{Bentley1975}. This data structure is useful in several applications involving multi-dimensional search keys (e.g., range searches and nearest neighbor searches). $k - d$ trees for tracking applications have been explored in several studies \cite{kim2010fast, buchanan2006interactive, gupta2016novel, saxena2020novel, pinkham2020quicknn}. While clustering approaches have been used in conjunction with tree type data structure in applications such as 3D SLAM and mapping, to the best of our knowledge this paper presents the first implementation of DBSCAN and $k-d$ tree to detect and track multiple indoor mobile robots for event cameras \cite{roa2019dyclee, ramesh2020low, ramesh2019dart, zihao2017event, ekseth2019optimized, nie2020research, shibla2018improving, stojkovic2020density}. 

An off-the-shelf RGB camera-based tracking system provides ground truth (with human corrections when necessary). The experiments featured two-treaded, differentially driven ground robots with an accelerometer, gyroscope, magnetometer, and encoder sensors onboard. The event camera used in this study was affixed to a stationary mount on the ceiling to provide a fixed frame of reference. When an event camera moves, the background suffers from clutter, making it difficult to distinguish the object of interest \cite{lakshmi2019neuromorphic}. 

A limitation of an event camera is its inability to detect a stationary object as no new events are generated. Another challenge of working with event cameras is the amount of background noise they can generate. In a multi-robot system, both these issues can cause spurious detections and loss of real-time tracks \cite{chen2020}. The $k - d$ tree-based tracking method presented here maintained robust tracks in the face of event camera noise and lack of events (due to robots moving slowly or stopping). This tracking method is named IDTrack. 

This study's main contributions are as follows: 
\begin{enumerate}
\item A density-based spatial clustering of applications with noise (DBSCAN) based real-time method to detect multiple ground robots operating in an indoor environment using a stationary event camera. 
\item A single k-dimensional ($k - d$) tree-based robust tracking technique called IDTrack to ensure that robot IDs are not lost/mislabeled during their operations due to background noise or lack of robot motion. 
\item Experiments including up to 4 robots to study the effect of i) varying DBSCAN parameters, ii) the event accumulation time, iii) the number of robots in the arena, and iv) the speed of the robots on the detection and tracking performance. The performance was evaluated using precision, recall, Mean Absolute Error, and Multi-Object Tracking Accuracy metrics \cite{Bernardin2007}.
\item Event-camera data, ground truth data, and key Python functionalities have been open-sourced for the benefit of the community \cite{RADVisionGit}. 
\end{enumerate}

The remainder of this paper is organized as follows. Section 2 discusses existing literature on the use of event cameras for robotic systems. In Section 3, the hardware and software architecture used in the experimentation is described. Section 4 provides a detailed discussion of the detection and tracking method developed in this study. Section 5 covers in-depth the results of the experiments. Section 6 concludes the paper and provides some future directions. 

\section{Related Work}
The use of event-based cameras continues to grow across a plethora of applications. Event-based cameras have been used in object/pedestrian tracking, surveillance and monitoring, and object/gesture recognition  \cite{delbruck2013robotic, glover2016event, litzenberger2006estimation, orchard2015hfirst, lee2014real, amir2017low, jiang2019mixed}. They have also been shown to be beneficial for depth estimation, structured light 3D scanning, visual odometery, optical flow estimation, HDR image reconstruction, and Simultaneous Localization and Mapping (SLAM)  \cite{rogister2011asynchronous, rebecq2018emvs, matsuda2015mc3d, Kueng2016, benosman2013event, zhu2018ev, cook2011interacting, kim2008simultaneous, kim2016real, rebecq2016evo, vidal2018ultimate}. For aerial robotics, safe navigation has been accomplished using the low perception latency afforded by event cameras \cite{falanga2019fast, dimitrova2020}. Readers interested in gaining an exhaustive understanding of event cameras are referred to the comprehensive survey paper \cite{gallego2020} and its references. 

\subsection{Robotic systems with event cameras}
Event camera-based algorithms for single or multiple object detection, pose estimation, and tracking (MOT) can be classified into three categories: feature-based, artificial neural network-based, and time surface-based \cite{lakshmi2019neuromorphic}. Studies focusing on robot pose estimation using event cameras have been reported in the literature \cite{bryner2019event, gallego2017event, muggler2017event}. In \cite{bryner2019event}, the authors validate a method to estimate the 6-DOF pose of a iniVation Dynamic and Active-pixel Vision Sensor DAVIS346 event camera given a photometric 3D map of the scene and improve upon the results of a similar study \cite{gallego2017event}. In \cite{muggler2017event}, a DAVIS camera-based approach was successfully evaluated for tracking a quadrotor motion performing high-speed maneuvers like flips with rotational speeds up to $1200^{\circ}/sec$. In \cite{ni2012asynchronous}, the authors validated an event-based iterative closest point (EICP) algorithm to estimate pose and track microgripper position at a frequency of 4 kHz using a DVS camera.

\subsection{Event camera-based robotic systems control}
One of the earliest applications of event cameras for feedback controls was the pencil balancing platform presented in \cite{conradt2009pencil}. Since then, several studies have shown how event cameras can be used for control of unmanned aerial systems, ground robots, robotic arms, and industrial robotic platforms \cite{dimitrova2020,falanga2019fast,conradt2009pencil, moeys2016steering, rueckauer2016evaluation, barrios2018movement, censi2013low}. In the field of unmanned aerial systems, event cameras are a viable solution for feedback control and optical flow  \cite{rueckauer2016evaluation, dimitrova2020, mueggler2014event, falanga2019fast}. In \cite{dimitrova2020}, the authors showcase an event-based feedback control for a quadrotor. The approach was evaluated on a dual copter platform for one-dimensional attitude control. In \cite{falanga2019fast}, the authors proposed dynamic obstacle avoidance for quadrotors using an event camera. The approach was evaluated in outdoor experiments where the quadrotor was capable of avoiding the obstacles moving at relative speeds up to $10$ meters/second. In \cite{mueggler2014event}, a DAVIS camera-based approach was successfully evaluated for tracking a quadrotor in motion performing high-speed maneuvers like flips with rotational speeds up to $1200^{\circ}/sec$. In \cite{rueckauer2016evaluation}, the authors compared nine optical flow algorithms that used events generated from a dynamic vision sensor using event cameras. The study highlighted the problems faced by standard optical flow algorithms such as Lucas-Kanade and local plane fit due to the noise in the event data stream and motion discontinuities. Event-based cameras in industrial robotics have recently become an active field of research \cite{barrios2018movement}. In \cite{barrios2018movement}, the authors compared the performance of a 2-axis servo-controlled robot based on the data acquisition and motion tracking from an event camera and a frame-based camera. The results showed that the robotic arm using the event-based camera could follow the object using image recognition while achieving up to $85\%$ percent data reduction and providing an average of $99$ms faster position detection than the frame-based camera. 

\subsection{Multi-object detection and tracking using event cameras}
A relatively small yet growing body of work underscores the value of event-based cameras for multi-object detection and tracking \cite{rodriguez2020asynchronous,ramesh2018long, mitrokhin2018event, liu2016combined, chen2019}.

 In \cite{rodriguez2020asynchronous}, the authors validated an approach for monitoring intruders using a DAVIS 346 DVS attached to a DJI Flamewheel F550 hexarotor. The approach included techniques to differentiate moving objects from the static objects in the moving background and had low computational cost. The scheme was implemented in the robot operating system (ROS) and validated using an unmanned aerial system for experiments performed in complex and unstructured scenarios during day and night.  In \cite{ramesh2018long}, the authors presented an approach for long-term tracking of objects using event cameras, even if the detected object left the scene and reappeared later. This method used an event-based local sliding window technique that performed reliably in scenes with a cluttered and textured background. In \cite{mitrokhin2018event}, the authors presented an approach for moving object detection and tracking using event cameras, which used information about the dynamic component of the event stream. The 3D geometry of the event stream was approximated with a parametric model to motion-compensate for the camera. Moving objects that did not conform to the model were detected in an iterative process. In \cite{liu2016combined}, the authors proposed an approach for object tracking that leveraged both frame-based and event-based camera sensors. The tracking algorithm was based on a conventional Convolutional Neural Network (CNN) based tracker combined with regions-of-interests from a cluster-based DVS tracker. The tracking system was evaluated on the Ulster dataset to solve the task of tracking an object of interest in a cluttered background with ego-motion. The results showed $90\%$ tracking accuracy with 20 pixel precision for the Ulster dataset. In \cite{chen2019}, the authors validated a pedestrian detector system based on multi-cue event information fusion. The system leveraged three different event-stream encoding methods -- Frequency, Surface of Active Event (SAE), and Leaky Integrate-and-Fire (LIF).

\begin{figure}[b!]
    \centering
        \includegraphics[scale = 0.43]{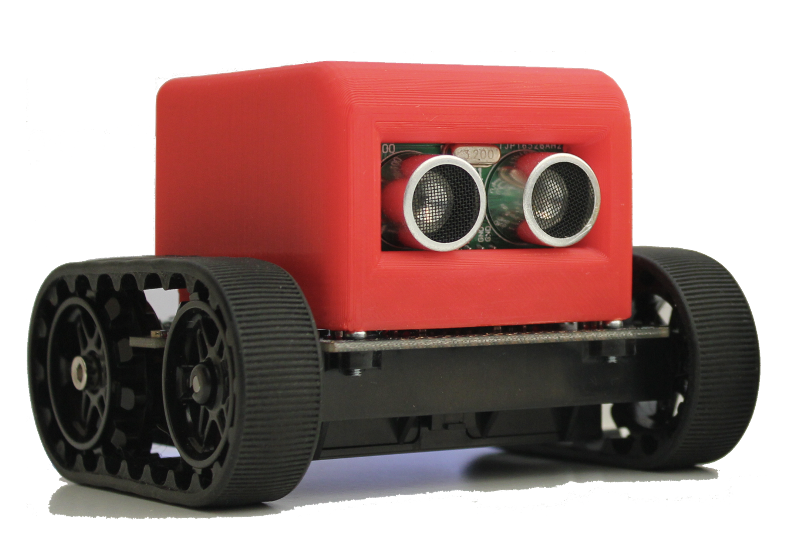}    
    \caption{The ground robot used in this study consisted of two treads that were differentially driven. The robot featured a 9-degrees of freedom inertial measurement unit (IMU) with an accelerometer, gyroscope, and magnetometer. The robot also had motor encoders that measured the drive motor position and rotational speed.}
    \label{robotfig} 
\end{figure}

 \begin{figure*}[t!]
    \centering
    \includegraphics[scale = 0.4]{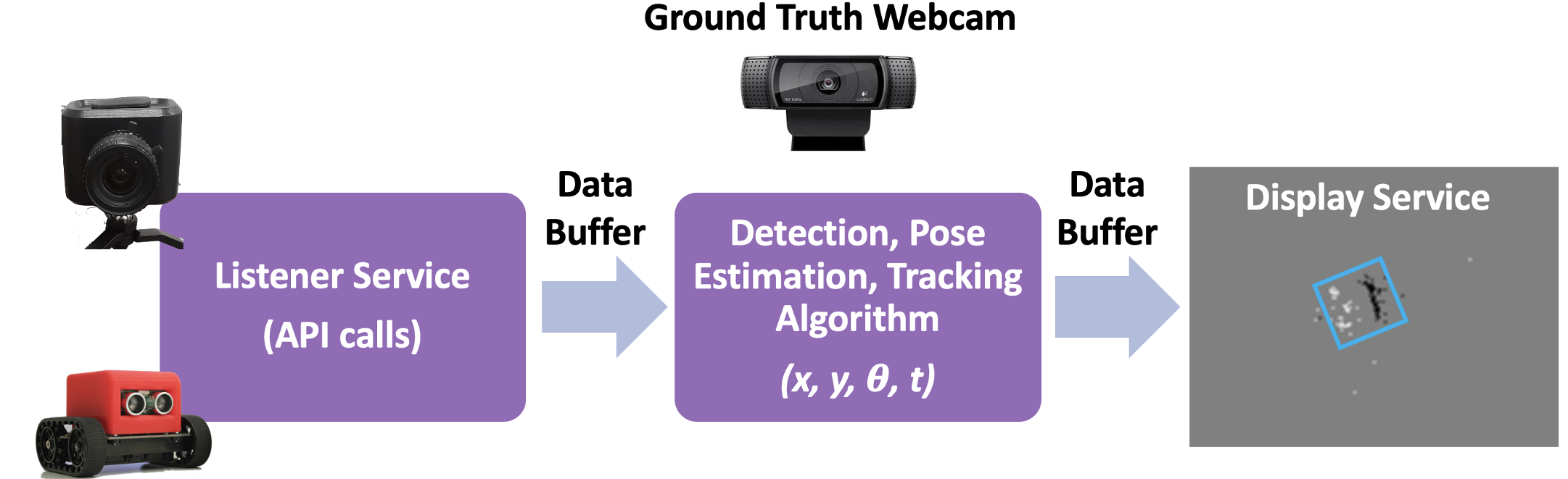}
    \caption{The C++/Python software application handled 1) data acquisition, 2) detection and tracking, and 3) data visualization in real-time. Two data buffers were used -- one for accumulating event data for a specified amount of time $t_{a}$ and the other for buffering data for annotated visualizations.}
    \label{fig:module_display} 
\end{figure*}
 
Closest to the technique presented in this paper is clustering based object detection and tracking explored in \cite{Bolten2019} for a single object and in \cite{Foster2019} for multiple objects. In \cite{Bolten2019}, the authors used event cameras to preserve a pedestrian's privacy while detecting his/her presence. The authors validate a proof-of-concept approach to cluster a single human in the cluttered environment, calculate the cluster centroid and track it over time. In \cite{Foster2019}, the authors presented a multi-object tracking technique that pre-filtered event data to reduce computational complexity, identified event clusters (representing multiple objects) using spatial variance. They tracked the identified clusters using a partial update Gaussian Mixture Probability Hypothesis Density (GMPHD) filter. The authors tested their approach on a simulated dataset only. The simulated dataset featured multiple virtual small Unmanned Aerial Vehicles (sUAVs) created using Blender 3D design software \cite{blender2021}. 

This paper extends this growing body of work by implementing a DBSCAN and $k-d$ tree-based approach to experimentally validate real-time detection and tracking of up to 4 ground robots operating in an indoor environment. 

\section{System Architecture} 
This section elucidates the hardware and software setup for capturing real-time event information from the camera.

\subsection{Hardware}
The hardware consisted of the event camera, an RGB webcam, and the ground robots. 

\begin{figure}[b!]
    \centering
    \includegraphics[scale = 0.22]{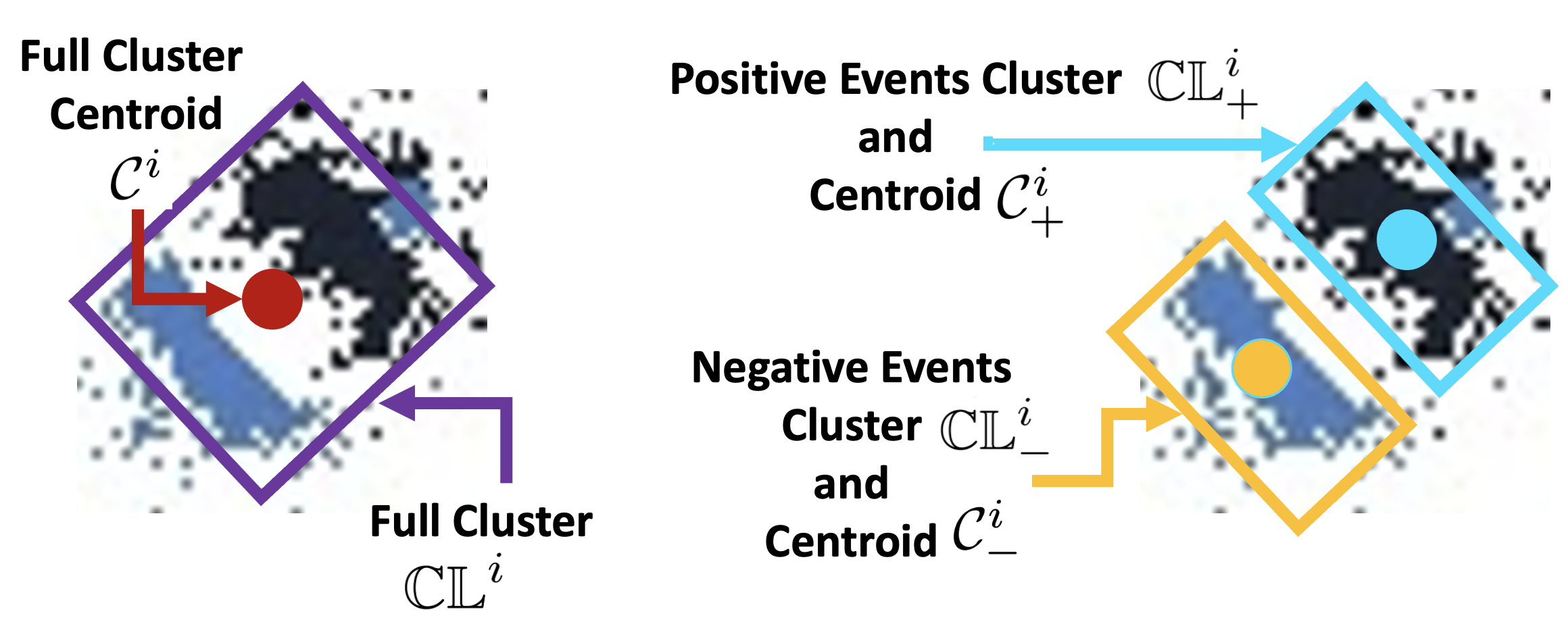}
    \caption{Positive event pixel locations were displayed using black color, negative event pixel locations were displayed using blue color, and pixels with no events were displayed using white color. Also shown here are the full robot cluster $\mathbb{CL}^i$ corresponding to the robot $i$ with centroid $\mathcal{C}^{i}$, the positive events cluster $\mathbb{CL}^i_{+}$ with centroid $\mathcal{C}^{i}_{+}$, and the negative events cluster $\mathbb{CL}^i_{-}$ with centroid $\mathcal{C}^{i}_{-}$.}
    \label{clusters} 
\end{figure}

\subsubsection{Camera specifications}
The event camera used in the experiments was a VGA-resolution contrast-detection vision sensor from Prophesee, shown in Figure \ref{hero_image}. This camera features a CMOS vision sensor with a resolution of 640x480 (VGA) pixels with $15\mu m \times 15\mu m$ event-based pixels and a high dynamic range (HDR) beyond 120 dB. The camera ran on 1.8V supplied via USB, with a 10$mW$ power dissipation rating in low power mode. The camera was interfaced using USB for communication and was mounted on the ceiling of the experiment area looking down. A Logitech C920 HD PRO webcam was also mounted next to the event-camera to capture the mission and provide ground truth measurements. Both cameras are shown in Fig. \ref{hero_image}. 
 
\subsubsection{Robot specifications}
The ground robot used in this study was a tracked robot based on the Arduino-compatible ATmega32U4 MCU and is depicted in Fig. \ref{robotfig}. It featured two 150:1 high-powered micro-metal gear motors with integrated dual motor drivers, a ring of RGB LEDs, quadrature encoders, accelerometer, gyroscope, and magnetometer. At 100\% motor power, the robots moved at approximately 0.46 m/s (and approximately 0.23 m/s @ 50\% motor power). The ground robots were networked using a Bluetooth connection and were programmed using a Python API.

\begin{figure*}[t!]
    \centering
    \includegraphics[scale = 0.6]{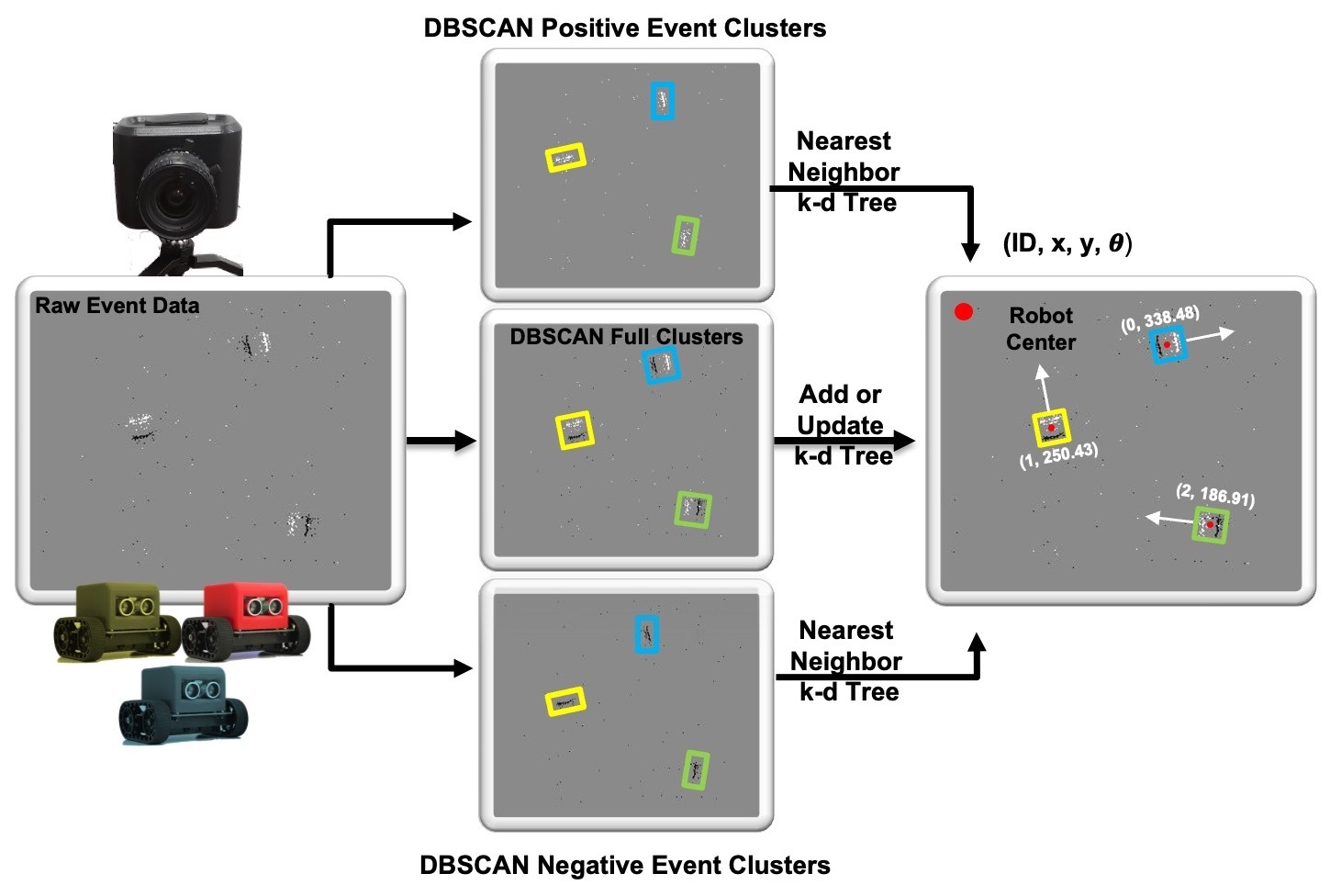}
    \caption{Robot detection and IDTrack using full, positive, and negative event clusters.}
    \label{visualization} 
\end{figure*}

\subsection{Software}
All software was developed in C++ and Python according to a modular architecture shown in Figure \ref{fig:module_display}. 
 
\subsubsection{Listener service}
The listener service used a set of data acquisition functionalities (API calls) provided by Prophesee to read event data. To reduce noise and decrease event data processing time, manufacturer-recommended parameter tuning was performed. The parameters set the operating point of the photoreceptor feedback amplifier, the bandwidth of the post-photoreceptor source follower buffer stage, the refractory period between events, contrast sensitivity, and high pass filtering \cite{biasSettings2021}. The event data contained information about the x, y location of a pixel where the event occurred and the event type (positive or negative). An event was positive when there was a positive change in light intensity, and it was negative when there was a negative change in light intensity. A data buffer continuously collected event objects from the event listener service. This information was then encoded into address-events that were asynchronously transmitted at the periphery via a mechanism called Address-Event Representation (AER). This process was repeated up to 30 times a second. The parameter accumulation time $t_{a}$ was used to specify the time in microseconds for which events were fetched from the past to the current time. These events were then stored in the data buffer before being passed to the detection and tracking service.

\subsubsection{Detection and tracking service}
The detection and tracking service featured a set of functionalities that ingested the event data from the AER data buffer and provided real-time ($x$, $y$) location and heading angle $\theta$ of each moving robot in the scene. This service was run 24 times per second to create a smooth track for each moving robot. The robots' detected location along with their respective IDs, were stored in a data buffer to be passed onto the display service. 

\subsubsection{Display service} 
A display service consisted of visualization functionalities with a variable display frame rate and a fixed frame matrix (640 pixels X 480 pixels). The frame rate specified the number of times the display service was run in one second.  The final image was created by reading every event in the display buffer and assigning an RGB value of (0, 0, 0) or (255, 255, 255) to the pixel location where the event occurred.  If the event was positive, the pixel at that location was assigned black color. If the event was negative, the pixel at that location was assigned white color. Image pixels that remained the same were assigned gray color. Additional visualizations were created (to add clarity for the readers) with black-colored positive events, blue-colored negative events, and white color for unchanged pixels as depicted in Figure \ref{clusters}. 

\subsubsection{Homography Service}
As the two cameras above the indoor testing arena were mounted next to each other, homography transformations were necessary to align their reference frames for accurate comparison \cite{ni2012asynchronous}. A homography step was performed at start-up to account for the distance and angle between camera lenses. Pixel coordinates of the four corners of the foam mat were noted from both the RGB and event cameras. A homography matrix was constructed from these eight 2D points, mapping corresponding corner locations. Finally, a perspective transform was performed on the RGB camera's captured image such that the event and RGB coordinate planes were aligned. The homography service was run at the beginning of each experiment session.

\section{Detection and Tracking}
Detection and tracking of multiple robots was performed using DBSCAN to create positive and negative event clusters and a single $k -$ dimensional ($k - d$) tree to keep track of robot location in the arena. Considering $i = 1, \ldots, n$ robots in the experimental arena, the following discussion elucidates the process depicted in Figure \ref{visualization}. 

\subsection{Detection: Position $(x,y)$ and Heading Angle $\theta$ estimation } 

The event-objects stored in the data buffer from the Listener service were used to create three arrays. The first array contained all the positive events in the data buffer, the second array contained all the negative events, and the third array contained all events. Each array was passed through the DBSCAN algorithm with $eps$ and $minPts$ values as critical parameters. The DBSCAN algorithm returned clusters of partial (positive and negative events) and full events. For each
full cluster $\mathbb{CL}^i$ corresponding to robot $i$, the $(\mathcal{C}^{i}x, \mathcal{C}^{i}y)$ coordinates of the cluster centroid $\mathcal{C}$$^i$  were determined by calculating the mean of $(x^{j}, y^{j})$ coordinates all the points $j$ in that cluster. 

\begin{equation}
(\mathcal{C}^{i}x, \mathcal{C}^{i}y) =  \displaystyle \left ( \displaystyle \frac{1}{j}\sum_{k=1}^{k=j} x^{k},  \displaystyle \frac{1}{j}\sum_{k=1}^{k=j} y^{k} \displaystyle \right ), \forall {i} \in {1\ldots n}, \forall {j} \in \mathbb{CL}^i
\end{equation}

The calculated $(x_{i}, y_{i})$ locations for each centroid $\mathcal{C}$$^i$ of the full clusters $\mathbb{CL}^i$ were then used to create robot objects with empty positive event cluster $\mathbb{CL}^i_{+}$ with centroid $\mathcal{C}^i_{+}$ and negative event cluster $\mathbb{CL}^i_{-}$ with centroid $\mathcal{C}^i_{-}$, and added to a 2-dimensional $k - d$ tree $\uptau$. Figure \ref{clusters} depicts an example of these clusters. $\mathcal{C}$, $\mathcal{C}_{+}$, and $\mathcal{C}_{-}$ represent the set of all full, positive, and negative event cluster centroids, respectively, for each timestep. 

\begin{algorithm}[b!]
\DontPrintSemicolon
 \KwData{$\mathcal{C}(t)$}
 \KwResult{$\uptau(t) \ni \{(\mathcal{C}^{i}x (t), \mathcal{C}^{i}y (t) )\}, \forall i \in \{1 \ldots n\}$}
 $kdtree\leftarrow$ emptyTree, $nextID$ = 1\;
 \While{{\mbox IDTrack} is running}{
            \For{$\mathcal{C}^{j}(t)$ in $\mathcal{C}(t), j \in \{1 \ldots n\}$ }{
        
        \eIf{$\uptau(t)$ is emptyTree}{
        $\uptau(t)\leftarrow$ addNode($\mathcal{C}^{j}x(t)$, $\mathcal{C}^{j}y(t)$, $nextID$)\;
        $nextID$++
        }{
        $(\mathcal{C}^{i}(t), {d}^{ij}(t))\leftarrow$ $\uptau(t) $.NN($\mathcal{C}^{j}x(t)$, $\mathcal{C}^{j}y(t)$)\;
        
        \eIf{${d}^{ij}$ $> \sigma$}{
        $\uptau(t)\leftarrow$ addNode($\mathcal{C}^{j}x(t)$, $\mathcal{C}^{j}y$(t), $nextID$)\;
        $nextID$++
        }{
        $\mathcal{C}^{i}(t)\leftarrow$ updateNode($\mathcal{C}^{j}(t)$)\;
        }
        }
     } 
 }
 \caption{IDTrack for Robust Tracking}
 \label{algo_kdtree}
\end{algorithm}

The ground robot's heading angle calculations from the event data relied on positive and negative event data captured from the event camera. The robot's heading angle was then calculated from the inverse tangent of the positive and negative cluster's centroid for each bot as described in \ref{eqn:headingAngle}. 

\begin{equation}\label{eqn:headingAngle}
\theta^{i} = \tan^{-1}\bigg(\frac{\mathcal{C}^{i}y_{-}-\mathcal{C}^i{y_+}}{\mathcal{C}^ix_{-}-\mathcal{C}^i{x_+}}\bigg), 
\end{equation} where $(\mathcal{C}^{i}x_{+},\mathcal{C}^i{y_+})$ are the coordinates of $\mathcal{C}^{i}_+$, and  $(\mathcal{C}^{i}x_{-},\mathcal{C}^i{y_{-}})$ are the coordinates of $\mathcal{C}^{i}_{-}$

In this manner, the location and heading angle of the robot $(\mathcal{C}^{i}x, \mathcal{C}^{i}y, \theta)$ were estimated. Figure \ref{visualization} depicts the resulting visualization with the frame of reference, bounding box, tracking id, and heading angle displayed on the image.

\subsection{IDTrack Technique for Robust Tracking} 
The detection and heading angle processes were repeated continuously for the entire duration of the mission $t = 0 \ldots T_{\mbox{mission}}$. At each time step $t$, the $k-d$ tree $\uptau(t)$ was updated with information about available robot clusters $\mathbb{CL}^i$  $\forall i \in \{1 \ldots n\}$ and their corresponding centroids $\mathcal{C}^{i}$ as shown in Algorithm \ref{algo_kdtree}.  

\begin{equation} \label{kdtree}
\uptau(t) \ni \{(\mathcal{C}^{i}x (t), \mathcal{C}^{i}y (t) )\}, \forall i \in \{1 \ldots n\}
\end{equation}

The tracking suffered from spurious noise effects that caused DBSCAN to assign new IDs to clusters. Additionally, when the robots would slow down or rotate in place, the event camera would report fewer events (and hence sparser clusters), thereby causing lost tracks/IDs or mislabeling of robots. To address these issues, IDTrack leveraged nearest neighbor searches between the robot cluster centroids. 

IDTrack was initiated as soon as information about the first full detected cluster centroid $\mathcal{C}^{i = 1}$ was added to the empty $k - d$ tree $\uptau(0)$.

After this initiation step, a nearest neighbor search $\uptau(t)$.NN was conducted for all other detected full event cluster centroids $\mathcal{C}^{j}(t)$ at a given time step $t$. The Euclidian distance $d^{ij}(t)$ between $\mathcal{C}^{j}(t)$ and its nearest neighbor $\mathcal{C}^{i}(t)\in \uptau(t)$, $\forall i, j \in \{1, \ldots n\}, i \ne j$ was used to define two possible cases:
\begin{enumerate}
\item {\bf New Robot Discovered:} If this distance $d^{ij}(t)$ was greater than the width of the physical robot chassis $\sigma$, the new cluster was inferred as a distinct robot. A new node corresponding to this newly discovered robot was created in $\uptau(t)$ with ($\mathcal{C}^{j}x(t), \mathcal{C}^{j}y(t)$) coordinates. A counter variable called $nextID$ was incremented by one each time a new node corresponding to a distinct robot was added to $\uptau(t)$. This counter helped keep track of the sequence of IDs being assigned to the newly discovered cluster centroids. 

\item {\bf Same Robot Rediscovered:} On the other hand, if $d^{ij}(t)$ was less than or equal to $\sigma$, the centroid $\mathcal{C}^{j}(t)$ was inferred to belong to the same robot represented by centroid $\mathcal{C}^{i}(t)  \in \uptau$. In this case, $\mathcal{C}^{i}(t)  \in \uptau$ was overwritten by $\mathcal{C}^{j}(t)$ as the latest centroid information about the corresponding robot. 
\end{enumerate}
 
Since the positive and negative event streams captured from the camera were not necessarily in the same order as the full event stream, additional data processing was performed to ensure that the positive and negative cluster centroids were assigned to the correct robot. This was achieved by running nearest neighbor searches between  $\mathcal{C}(t) \in \uptau(t)$ and $\mathcal{C}_{+}(t)$, and  $\mathcal{C}(t) \in \uptau(t)$ and $\mathcal{C}_{-}(t)$ at each time step $t$.  

\begin{figure}[t!]
    \centering
    \includegraphics[scale = 0.19]{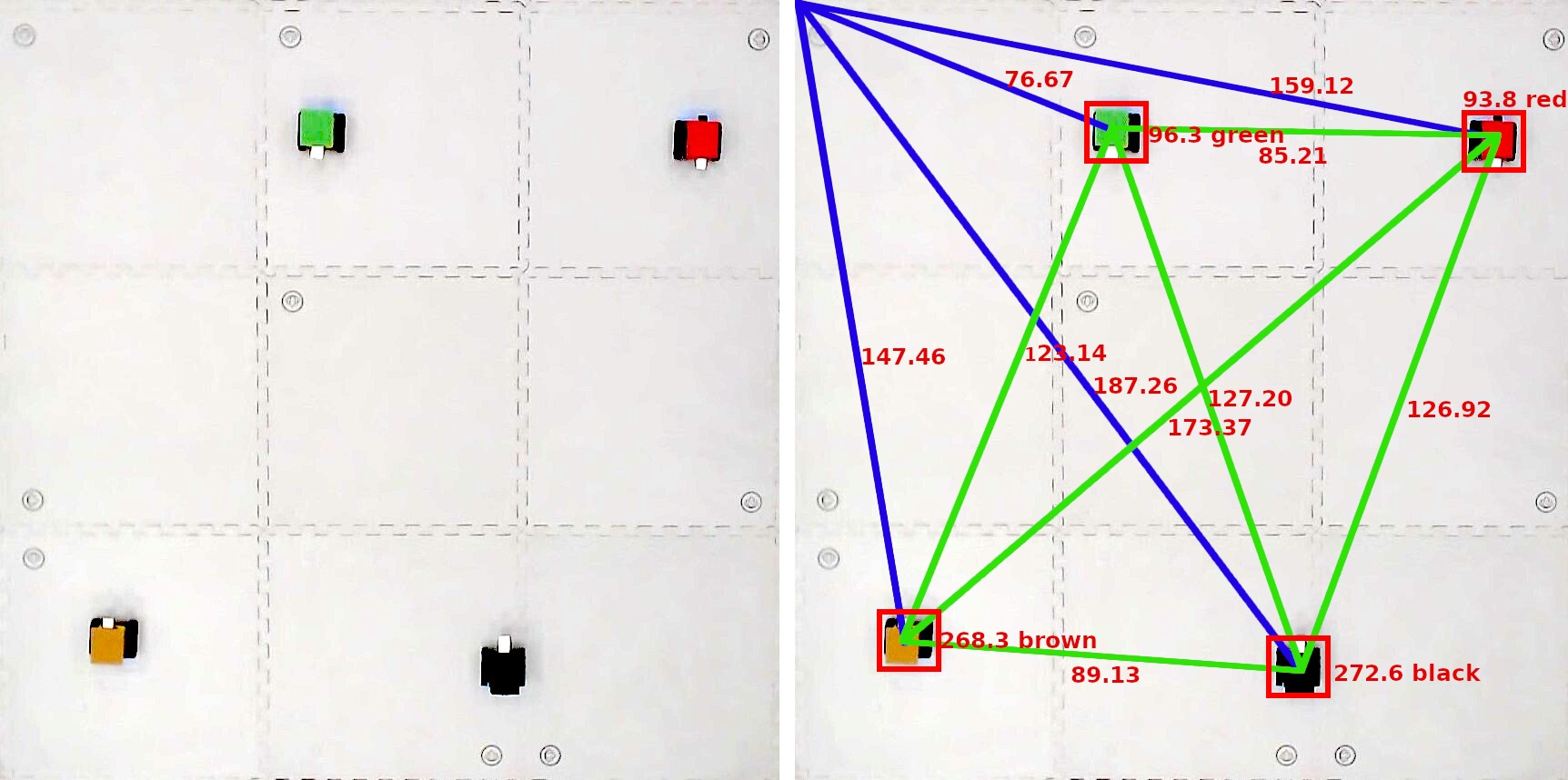}
    \caption{Left: Four robots in a RGB webcam frame. Right: RGB tracking information depicting distances and heading angles. }
    \label{RGBfigure} 
\end{figure}

\begin{figure}[b!]
    \centering
    \includegraphics[scale = .19]{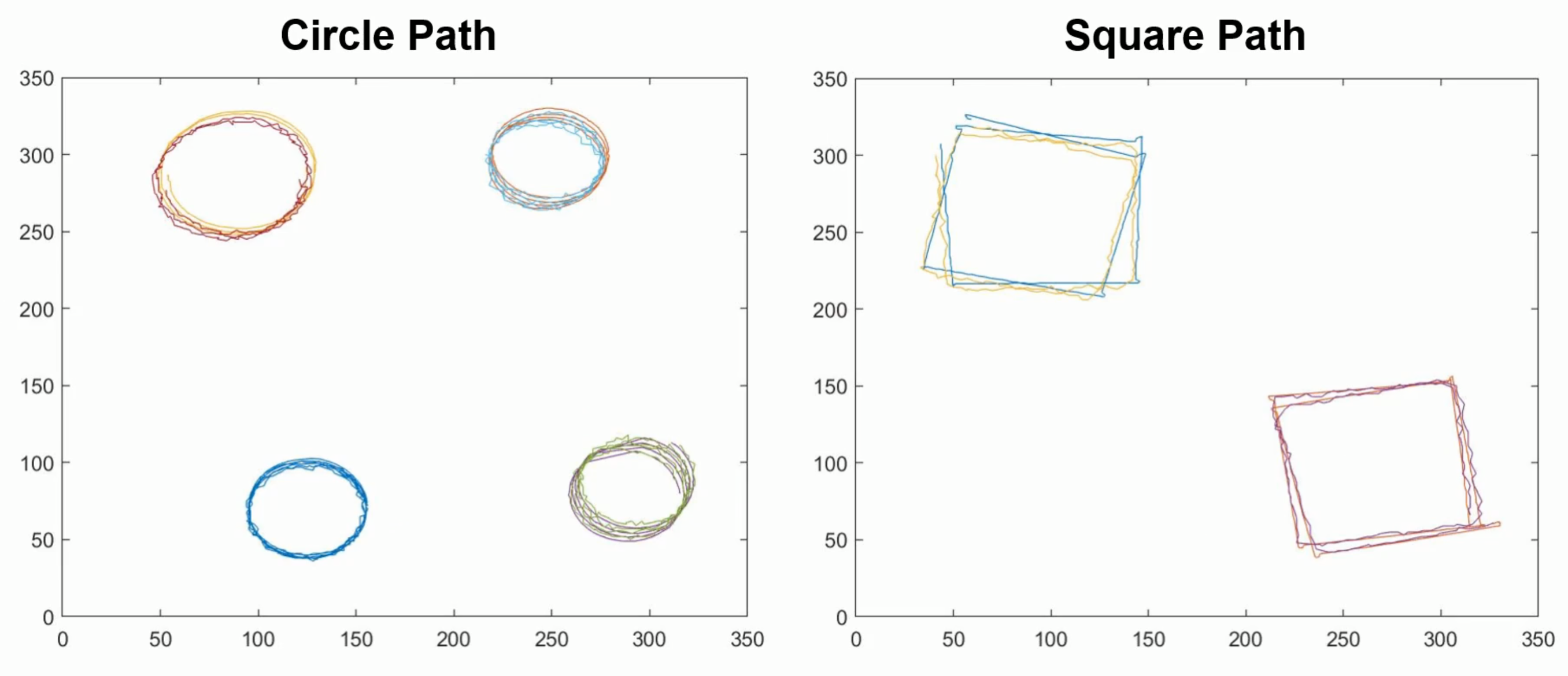}
    \caption{Left image: 4 robots circular paths. Right image: 2 robots square paths. The only feedback control applied to the robots was a PID controller using motor encoders. As such, the tracing of the shapes was not perfect, as seen here. }
    \label{shapes} 
\end{figure}

\subsection{RGB ground truth data}
A RGB webcam-based detection, robot heading angle estimation, and tracking system was developed in Python to provide ground truth data. An output frame of this system is depicted in Figure \ref{RGBfigure}. This was a frame-based system developed using OpenCV libraries to detect and identify robots based on the color of their 3D printed shell \cite{itseez2015opencv}. The algorithm provided a $k - d$ tree containing the centroid locations of each robot in the frame. The use of a $k - d$ tree along with OpenCV libraries has been studied in the literature \cite{choi2012fast, ragaglia2014multiple, shoaib2014hardware}. This system first converted the input RGB image to HSV representation. Using relevant OpenCV library functions such as {\tt findContours()}, the system was used to detect each robot's area in the frame and calculate the robot's centroid. This process was repeated at 24 frames per second to generate tracking information about the robots in the arena. These results were manually cross-checked and corrected for any labeling errors/missed detections for a high-quality ground truth dataset.

\section{Experimental setup and results}
The event-based camera and the webcam were mounted on the ceiling directly above a 183 cm $\times$ 183 cm area. The area was covered with white foam mats. At the start of each experiment, the robots were placed on the white foam mat, as depicted in Figure \ref{hero_image}. In all experiments, the robots were programmed to trace predefined paths (circle or square) on the mat via the Python interface, as shown in Figure \ref{shapes}. The event-based and RGB frame-based software trackers were simultaneously executed. The host system was configured with the Intel i7 8th generation processor @ 1.8 GHz and 16 GB RAM. The runs lasted between 15 seconds to 1 minute. 

\subsection{Key Metrics}
The key metrics used in the study are discussed next.  

\subsubsection{Detection Metrics}
The detection performance was assessed using Precision,  Recall, and Mean Absolute Error. 
Precision is the ratio of the number of correct detections to the total number of detections, recall is defined as the ratio of the number of correct detections to the total number of true objects in the data, and Mean Absolute Error is a measure of errors between paired measurements:
    \begin{equation}
        Precision = \frac{TP}{TP+FP}
    \end{equation}
    \begin{equation}
        Recall = \frac{TP}{TP+FN}
    \end{equation}

    \begin{equation}
        Mean\, Absolute\, Error = \frac{1}{n}{\displaystyle\sum_{i=1}^{n_{m}} |b_i - a_i|}
    \end{equation}
    where $TP$, $FP$, $FN$, $b_i$, $a_i$, and $n_{m}$ are the number of True Positives, False Positives, False Negatives, event data measurements, ground truth data measurements, and the number of measurements.
\subsubsection{Tracking Metric}
The tracking system performance was assessed using the multiple object tracking accuracy (MOTA) metric proposed by Bernardin \cite{Bernardin2007}. 
MOTA is defined as
\begin{equation}
    MOTA = 1-\frac{\sum_{k} M_k+FP_k+mme_k}{\sum_{k} g_k}
\end{equation}
where $M_k$, $FP_k$, and $mme_k$ represent the number of missed sequence, false positives, and mismatches in frame $k$, respectively. $g_k$ is the number of ground truth objects in frame $k$.

\begin{figure}[t!]
    \centering
    \includegraphics[scale = 0.8]{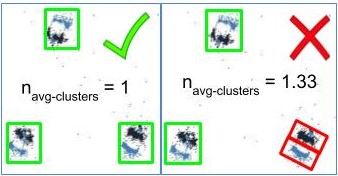}
    \caption{$n_{avg-clusters}$ metric calculations as applied to full clusters for DBSCAN for a given $minPts$ value. The image on the left depicts 3 correctly detected full event clusters with $n_{avg-clusters} = 1$. The image on the right depicts incorrect detection, which results in an $n_{avg-clusters} = 1.33$.}
    \label{n_c} 
\end{figure}

\begin{figure}[b!]
    \centering
    \includegraphics[scale = 0.26]{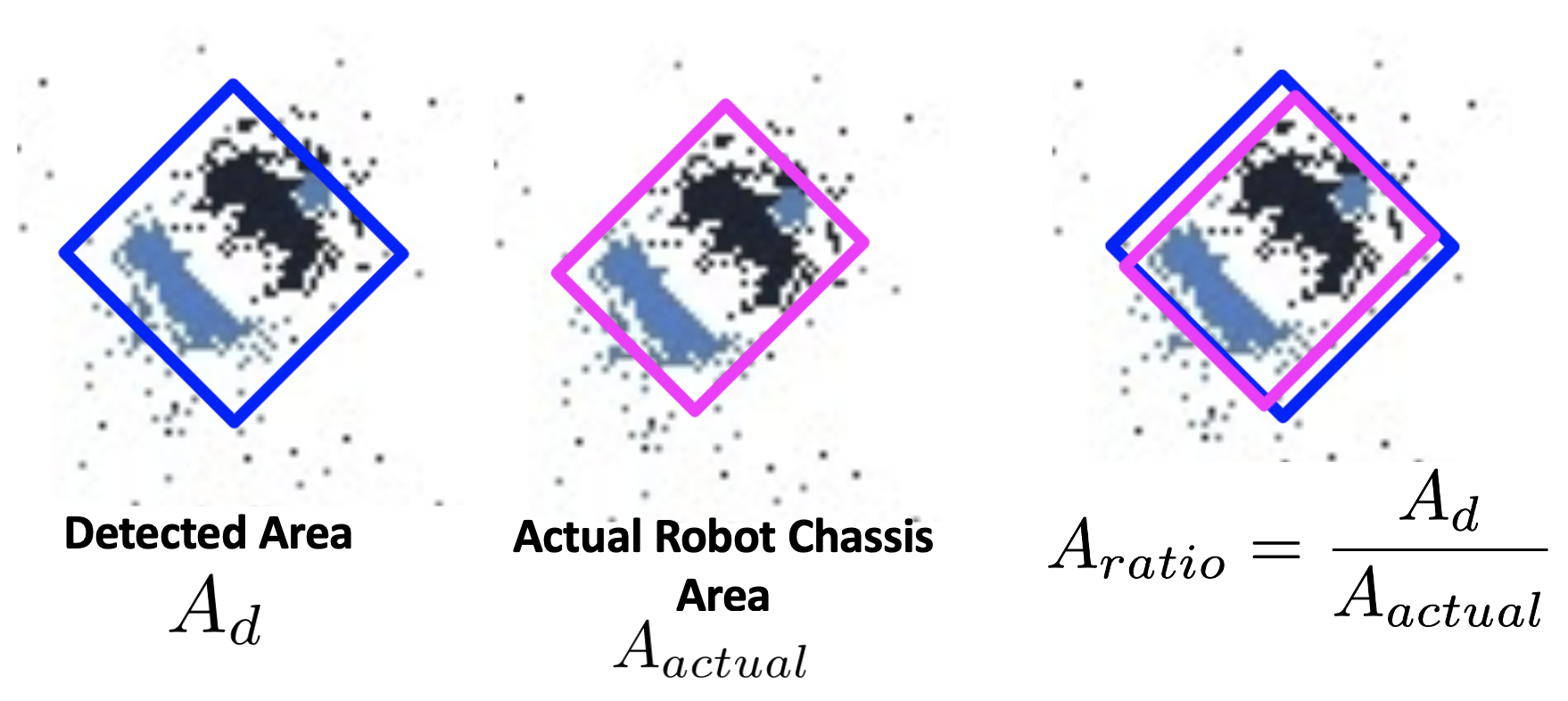}
    \caption{An example of $A_{ratio}$ metric calculations as applied to full event clusters for DBSCAN. The detected area $A_{d}$ is enclosed by the blue rectangle. The actual physical robot chassis area $A_{actual}$ is represented by pink rectangle. Ideally, $A_{ratio}$ should be as close to 1 as possible for a given $minPts$ value. ($t_{a}$ = 100,000 $\mu$s)}
    \label{a_ratio} 
\end{figure}

    \begin{figure*}[t!]
    \centering
    \includegraphics[scale = 0.33]{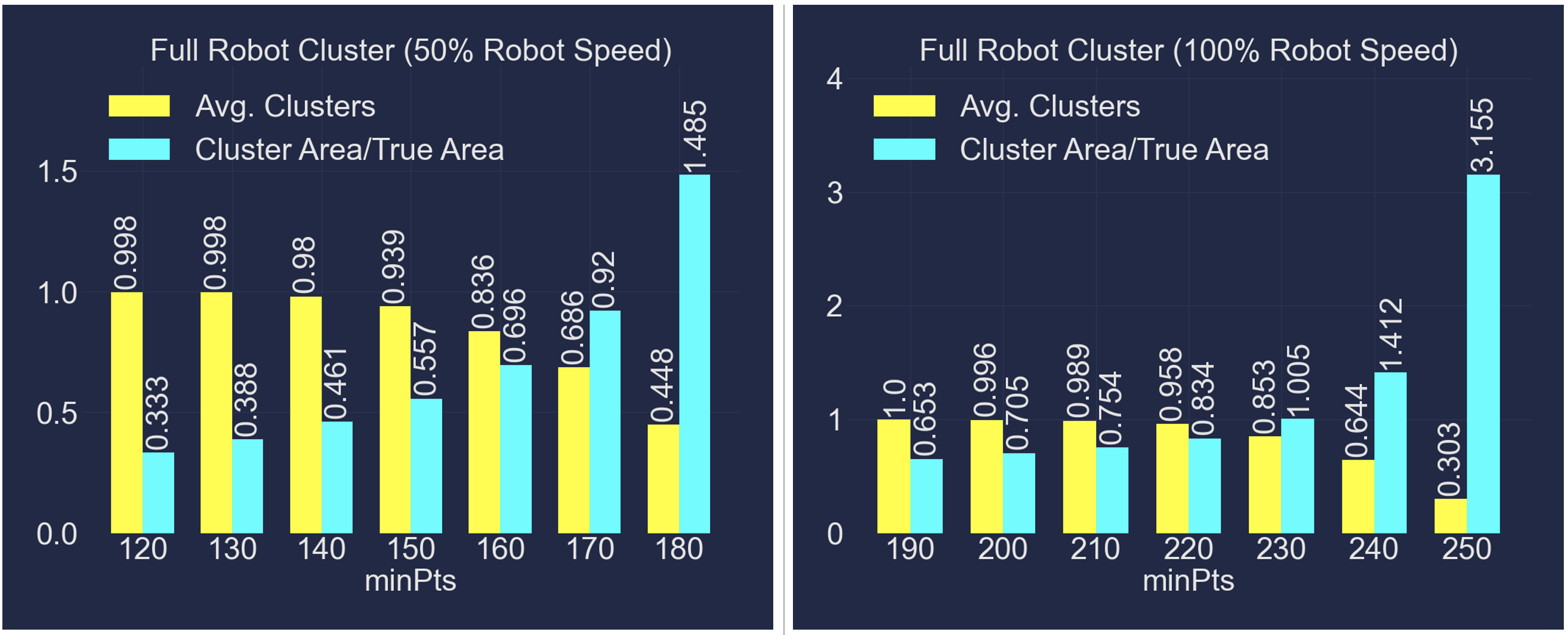}
        \includegraphics[scale = 0.34]{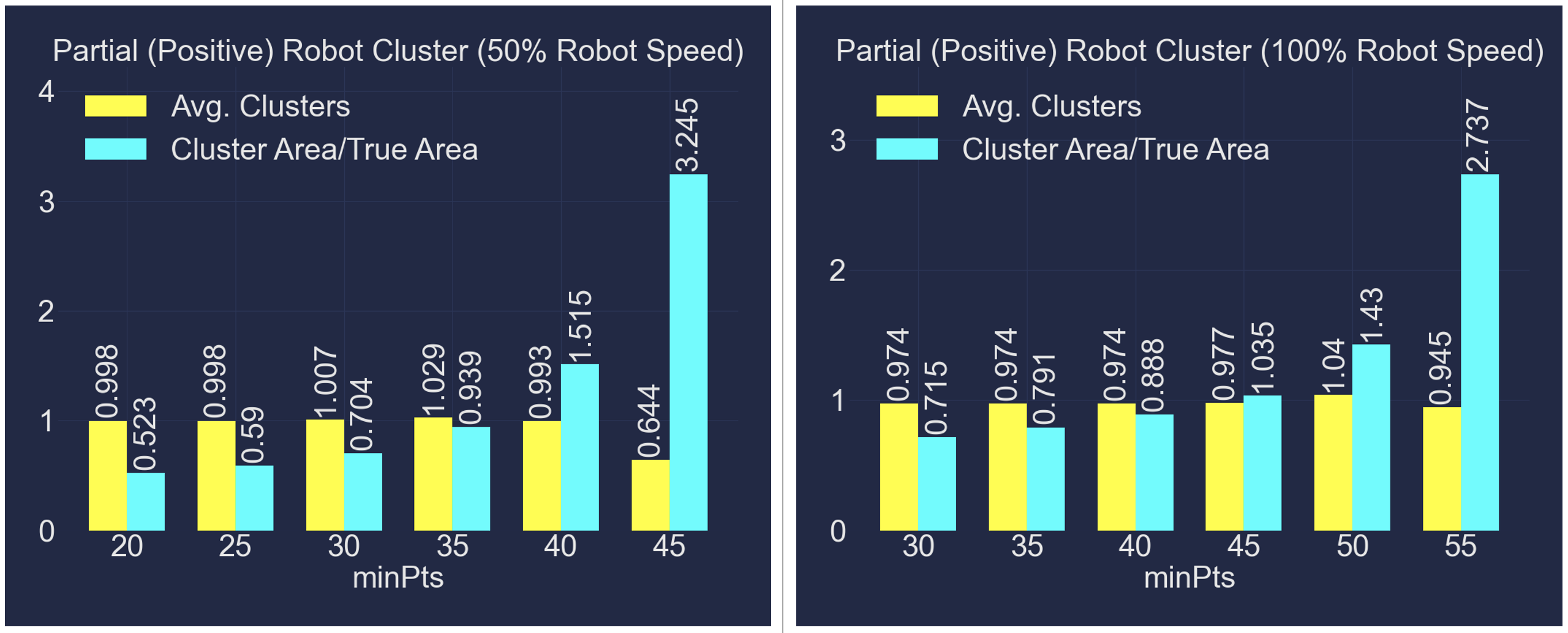}
    \caption{To understand the effect of $minPts$, the average number of clusters detected per robot ($n_{avg-clusters}$) and the ratio of detected cluster area and robot body area ($A_{ratio}$) were noted for a single robot for two scenarios (50\% motor power and 100\% motor power). The $minPts$ value (or range of values) that provided the values of $n_{avg-clusters}$ and $A_{ratio}$ closest to 1 were used for subsequent experiments involving more robots. ($t_{a}$ = 100,000 $\mu$s)}
    \label{minPts_figure} 
    
\end{figure*}

\subsubsection{Clustering Metrics}
The clustering performance was assessed using $n_{avg-clusters}$ and $A_{ratio}$. Robot area analysis was performed with $t_{a}$ = 100,000$\mu$s.
\begin{enumerate}
    \item $n_{avg-clusters}$: the average number of clusters detected per robot. Figure \ref{n_c} illustrates this metric.
    \begin{equation}
    n_{avg-clusters}= \displaystyle \frac{1}{n}{\sum_{k=1}^{n_{clusters}}\mathbb{CL}^{k}}
    \end{equation} where $n_{avg-clusters}$ is the number of clusters detected and $n$ is the number of robots. 
    \item $A_{ratio}$: the ratio of detected cluster area $A_{d}$, and actual robot chassis area $A_{actual}$. Figure \ref{a_ratio} illustrates for this metric.

 \begin{equation}
    A_{ratio}= \displaystyle \frac{A_d}{A_{actual}}
    \end{equation}
\end{enumerate}

\subsection{Experimental results}
The experiments focused on studying the detection and tracking performance of the proposed method in scenarios 1) with varying the DBSCAN $minPts$ parameter, 2) with changing $t_{a}$ of the camera, 3) involving 2, 3, and 4 robots, 4) with varying speed (at 50\% motor power and 100\% motor power) of the robots, and 5) variation in ambient light conditions. Each experiment was conducted 3 times and the average values over these 3 runs are noted in the following discussions.

\subsubsection{Effect of changing $minPts$ on robot detection}

The effect of changing the $minPts$ parameter value on the detection performance was evaluated using the metrics $n_{avg-clusters}$ and $A_{ratio}$. 

Figure \ref{minPts_figure} depicts the values of these two metrics averaged across 3 runs for a single robot moving at 50\% speed and 100\% speed, respectively. Ideally, for a single robot, DBSCAN should detect one full-body cluster on average per robot, $i.e.,$ $n_{avg-clusters}$ = 1. DBSCAN should also provide $A_{ratio}$ = 1. 

Ideally, for partial positive (or partial negative) clusters, DBSCAN should detect one partial positive (or partial negative) cluster on average per robot. The robot's partial cluster area was measured to be one-third of the overall area of the robot chassis. 

Figure \ref{minPts_figure} shows that at 100\% robot speed,  $minPts = 45$ provided the best values for $n_{avg-clusters}$ and $A_{ratio}$ for the partial positive clusters. For 50\% robot speed, this value was $minPts = 35$ for the partial positive clusters. The same value was used for the partial negative clusters. Similarly, a $minPts$ value between 220 and 230 provided the best $n_{avg-clusters}$ and $A_{ratio}$ values for the full robot clusters. The $minPts$ values noted here were used for all subsequent experiments.

The number of events captured by the event camera is affected by the speeds of the robots. This dependence affects the overall cluster quality, as observed in Figure \ref{minPts_figure}. Of note is the 50\% speed and full robot cluster scenario where changing $minPts$ between 150 and 170 dramatically affected $n_{avg-clusters}$ and $A_{ratio}$ values. 

{\bf Key Insight:}  $minPts$ is a critical parameter for the DBSCAN algorithm. As reported in Figure \ref{minPts_figure}, comprehensive tests can provide a $minPts$ value or range of $minPts$ values that lead to the best results for $n_{avg-clusters}$ and $A_{ratio}$.

\begin{figure}[b!]
    \centering
    \includegraphics[scale = 0.38]{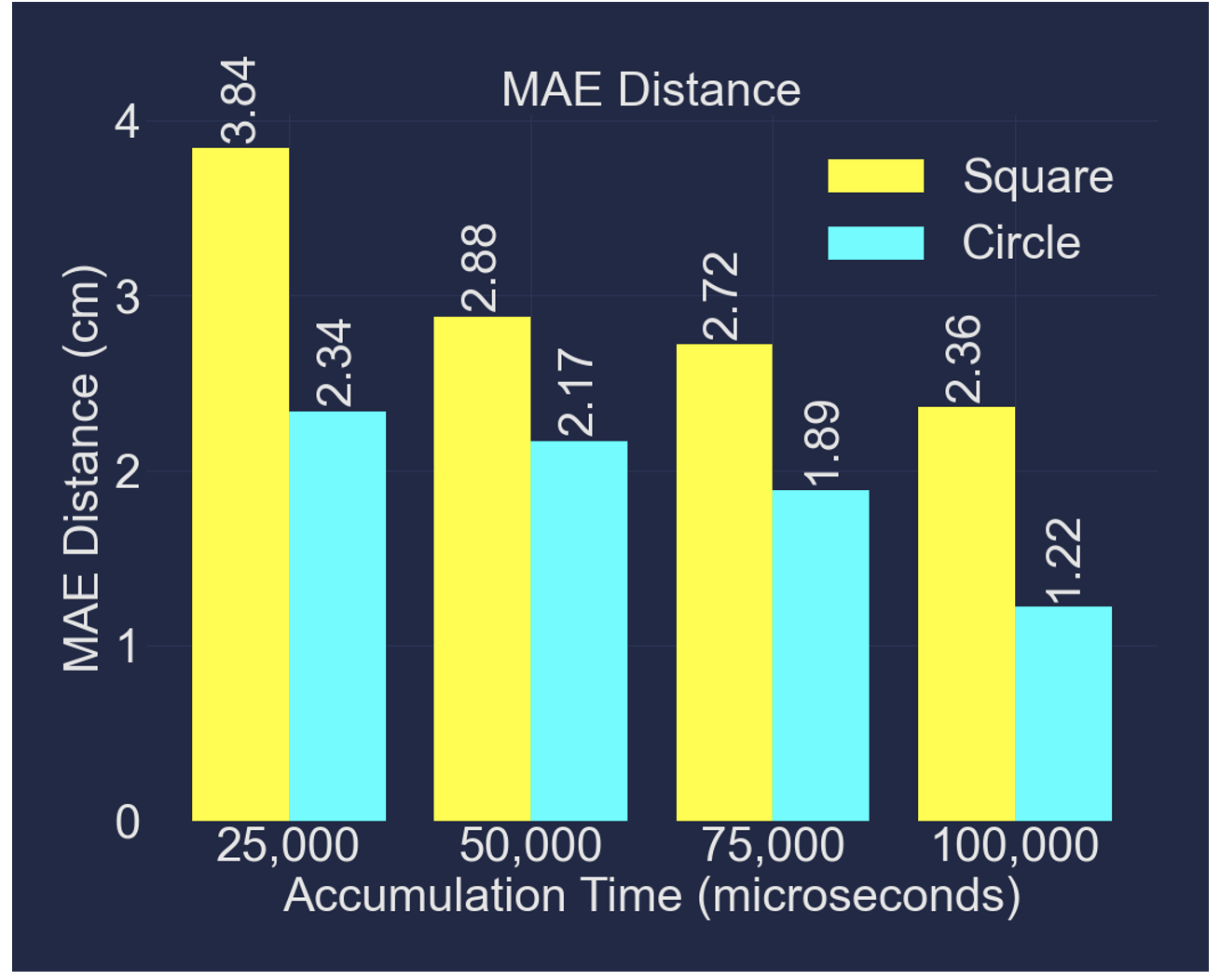}
    \caption{The effects of changing accumulation time $t_{a}$ on Mean Absolute Error (100\% motor power) for a single robot. $t_{a}$ = 100,000$\mu$s provided the lowest MAE results and was used for subsequent experiments.}
    \label{minPts} 
\end{figure}

\subsubsection{Effect of changing camera accumulation time $t_{a}$}
The effect of changing the $t_{a}$ value on detection performance was evaluated using the MAE distance metric. Figure \ref{minPts} depicts the values of this metric averaged across 3 runs for a single robot moving at 100\% speed for 4 different $t_{a}$ values. 

$t_{a}$ values of 100,000$\mu$s yielded better detection results than lower $t_{a}$ values in otherwise equivalent experiments. MAE distance results improved for circular and square patterns as $t_{a}$ increased. MAE results for circular paths were lower than square paths. This difference in the results between the two path patterns is attributed to the constant motion of circular paths, where the robot did not pause to turn, and detections were consistent. When the robot made $90^{\circ}$ zero-point turns for the square path pattern, fewer events were generated relative to when it traced the edges of the square. This reduced the detection quality and resulted in higher MAE. 

$t_{a}$ = 100,000$\mu$s was selected for all subsequent experiments.

{\bf Key Insight:} At a given robot speed, lower accumulation times led to fewer events buffered by the listener service. This caused sparser clusters and higher MAE. Accumulation time thus becomes a pivotal parameter to appropriately tune the sensitivity of an event camera to the change in brightness of objects in the field of view of the camera.

\subsubsection{Effect of increasing the number of robots}
The effect of increasing the number of robots on detection performance was evaluated using Precision, Recall, MAE distance, and MOTA metrics.

Table \ref{numbers} presents the values of these metrics averaged across 3 runs for scenarios with 1, 2, 3, and 4 robots. It is observed from Table \ref{numbers} that Precision, Recall, and MOTA remained uniformly high throughout most experimental runs, with a slight decrease observed in Recall and MOTA metrics during the 4 robot experiments. MAE distance increased as the number of robots in the experiment increased. For most cases, circular patterns reported the least MAE distance. 

{\bf Key Insight:} The robots were operating on white-colored foam mats. Different robot bodies generated varied numbers of events  depending on the color of their 3D printed shell. For example, the robot with a black-colored shell resulted in denser positive and negative event clusters compared to the robot with the yellow-colored shell. As the number of robots increased, a wider range of body colors was introduced into the experiments leading to an increase in MAE distance.

\begin{table}[t!]
\begin{center}
\caption{Performance metrics with an increasing number of robots in the arena. The performance metrics are listed for two different robot speeds (100\% motor power, 50\% motor power).}
\label{numbers}
\begin{tabular}{|l|l|l|l|l|l}
\hline
 & \multicolumn{3}{c|}{\bf{Detection}} & \bf{Tracking}   \\ \hline
 \hline
 $n$ & Precision & Recall & MAE distance (cm)   & MOTA 
 \\ \hline \hline
  \multicolumn{5}{c|}{Circle Pattern}     \\ \hline \hline 
1 & (1, 1) & (1, 1) & (1.22, 2.16)  & (1, 1)  \\ \hline
2 & (1, 1) & (1, 1) & (2.00, 2.82) & (1, 1)  \\ \hline
3 & (1, 1) & (1, 1) & (2.05, 3.78) &   (1, 1)   \\ \hline
4 & (1, 1) & (0.94, 0.93) & (4.32, 5.92)   & (0.94, 0.93)   \\ \hline
\hline
 \multicolumn{5}{c|}{Square Pattern}     \\ \hline
 \hline
1 & (1, 1) & (1, 1) & (2.36, 3.5)   & (1, 1)   \\ \hline
2 & (1, 1) & (1, 1) & (3.85, 3.2) & (1, 1)   \\ \hline
3 & (1, 1) & (1, 1) & (2.9, 4.96) &   (1, 1)   \\ \hline
4 & (1, 1) & (0.95, 1) & (4.31, 6.77)  & (0.95, 1)   \\ \hline
\end{tabular}
\end{center}
\end{table}

\subsubsection{Effect of changing motor speeds}
Motor speeds were changed by changing the power to the robot drive motors. The effect of changing motor speed value on detection performance is reported in Table \ref{numbers}. Key metrics used to evaluate the effect of motor speed on performance are Precision, Recall, MAE, and MOTA. Table \ref{numbers} reports the detection and tracking results of experiments with 1, 2, 3, and 4 robots, respectively. Two sets of experiments were conducted -- at 100\% motor power ($\sim$0.46 m/s robot speed) and at 50\% motor power ($\sim$0.23 m/s robot speed), respectively. 

It is observed that Precision, Recall, and MOTA metrics remained high regardless of motor speed in both path patterns. For square patterns, MAE increased as the robot number increased. Square patterns produced higher MAE compared to circular path patterns. At $50\%$ motor power, MAE distance was greater than at $100\%$ motor power with otherwise equivalent parameters. 

{\bf Key Insight}: Event cameras report events as per-pixel brightness changes. Slow-moving robots or robots that stop moving create less dramatic changes in brightness (and hence sparser clusters) than robots moving faster. Detection on slow-moving objects, therefore, leads to higher MAE. 

\begin{figure*}[t]
    \centering
    \includegraphics[scale = .5]{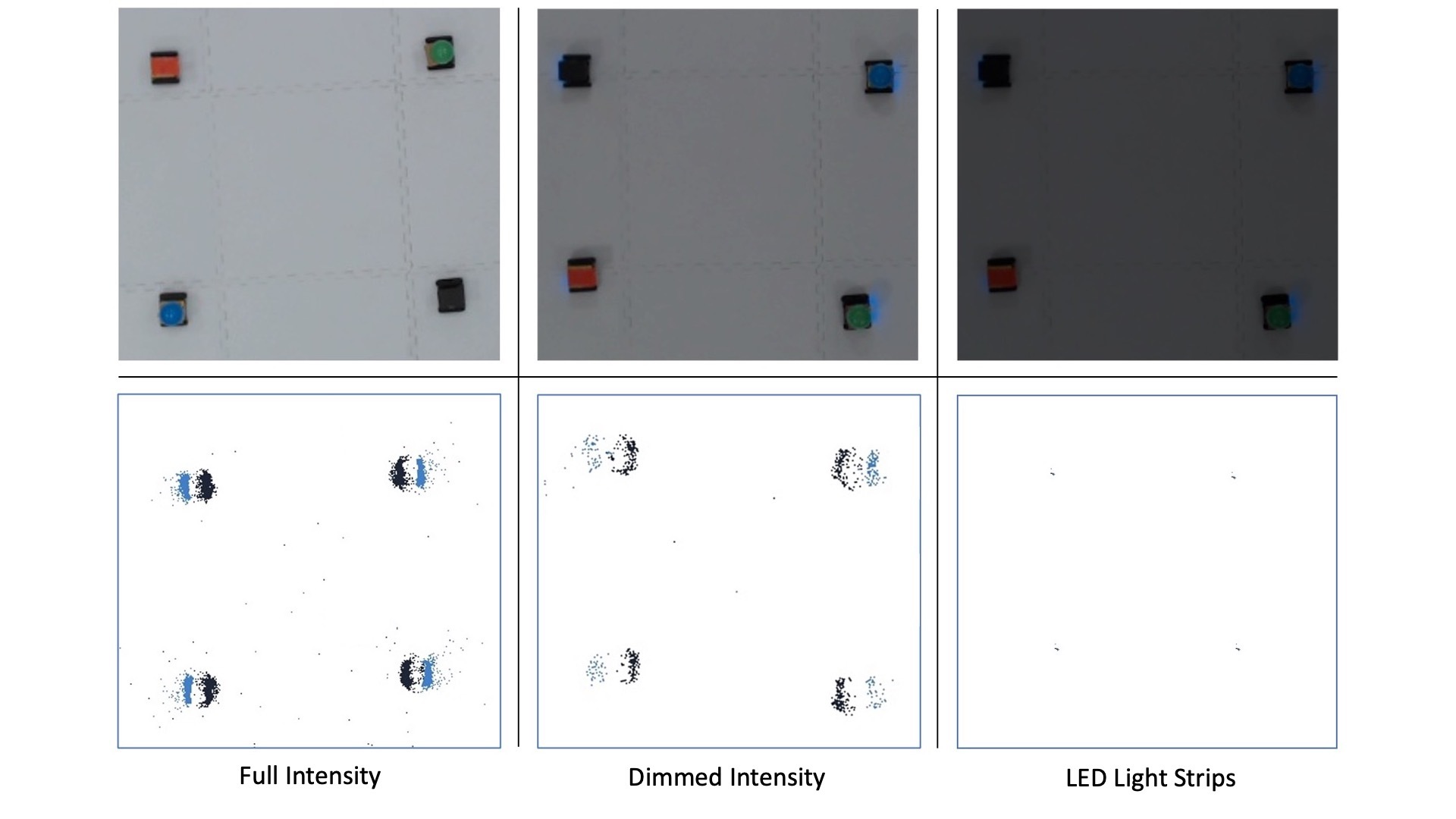}
    \caption{Various ambient lighting conditions shown for top-down four-robot scenarios. Top row: RGB camera. Bottom row: Event camera.}
    \label{brightness} 
\end{figure*}

\subsubsection{Effect of varying ambient lighting conditions}
Three ambient lighting settings were created by using lighting dimmers and LED light strips as depicted in Figure \ref{brightness}. The three conditions featured fluorescent lights at full intensity, fluorescent lights at dimmed intensity, and the use of LED light strips. Key metrics used to evaluate the effect ambient lighting conditions on performance are Precision, Recall, MAE, and MOTA. A graceful degradation in these metrics was observed as the ambient lighting was modulated from the brightest to the darkest settings. The Precision, Recall, MAE, and MOTA metrics for full brightness are noted in Table \ref{numbers}. For dimmed brightness, the Precision, Recall, MAE, and MOTA degraded to (1, 1, 3.01, 1) respectively for 1 robot, (1, 1, 2.59, 1) respectively for 2 robots, (1, 1, 3.12, 1) respectively for 3 robots, and (1, 0.73, 10.05, 0.72) respectively for 4 robots. All robots were commanded to move in square path patterns. Finally, for the darkest condition, the event camera struggled to detect motion consistently. 

{\bf Key Insight}: While event cameras can operate in varying lighting conditions, their performance is dependent on the overall ambient light intensity. As such, environmental lighting conditions should be considered while evaluating the detection and tracking performance of event camera based systems. 

\subsection{A note about heading angle calculations}
Heading angle calculations were performed using the positive and negative cluster centroids as described in Eqn. \ref{eqn:headingAngle}. For a single robot moving in a circular pattern, the minimum MAE $\theta$ recorded was $5.76^\circ$,  and the maximum MAE $\theta$ recorded was $13.01^\circ$. By contrast, the minimum MAE $\theta$ for a single robot moving in a square pattern was $30.52^\circ$, and the maximum MAE $\theta$ was $50.56^\circ$. A similar trend for MAE $\theta$ was observed for the multi-robot case. The following two key reasons contribute to the MAE $\theta$ results: 
\begin{enumerate}
    \item As mentioned earlier, during the $90^{\circ}$ zero-point turns for the square path pattern, the number of events generated was significantly less relative to when the robot traced the edges of the square. This reduced the detection quality and resulted in higher MAE. 
    \item Additionally, the color of the 3D printed shell of the robot also affected the number of events generated (and hence MAE $\theta$). 
\end{enumerate}
Further reductions in MAE $\theta$ may require use of probabilistic or optical flow techniques -- this is a topic of further investigation \cite{gallego2020}.

\begin{figure}[t!]
    \centering
    \includegraphics[scale = 1]{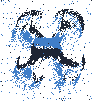}
    \caption{In-flight quadrotor visualized as pixel events. Future work will focus on detecting and tracking concurrent moving quadrotors.}
    \label{eventDrone} 
\end{figure}

\section{ Conclusion}
This study presented a method to detect and track mobile indoor ground robots using event cameras. Using DBSCAN and $k-d$ trees, this method achieved comparable performance to existing frame-based detection and tracking methods without the need for any training. With high detection and tracking fidelity in the face of event camera noise and robots stopping, experimental evaluations point to this method's suitability for real-time robot control applications. Future work will aim to extend this study to detection/track multiple quadrotors as displayed in Figure \ref{eventDrone}.

\bibliographystyle{IEEEtran}
\bibliography{ref}
\end{document}